\newif\ifisfinal\isfinaltrue
\documentclass[11pt,a4paper]{article}

\usepackage{dblfloatfix}
\usepackage{amsmath}
\usepackage{zllist}
\usepackage{cpbert}

\cpbertdocvar%

\usepackage{zvacl}
\zaveninit%

\zzaddauthors%

\newcommand{\tabsize}{\footnotesize}
\newcommand{\smalltabsize}{\tabsize}

\begin{document}
\zavenbegindoc%

\begin{abstract}
  \begin{nohyphenation}
  Many models are pretrained on redacted text for privacy reasons.  Clinical
  foundation models are often trained on de-identified text, which uses special
  syntax (masked) text in place of \acl{hphi}.  Even though these models have
  increased in popularity, there has been little effort in understanding the
  effects of training them on redacted text.  In this work, we pretrain several
  encoder-only models on a dataset that contains redacted text and a version
  with replaced realistic pseudo text.  We then fine-tuned models for the
  \acl{hphi} de-identification task and show how our methods significantly
  outperform previous baselines.  The contributions of this work include:
  \begin{zlenumerateinline}
  \item our novel, and yet surprising findings with training recommendations
  \item redacted text replacements used to produce the \pds%
  \item pretrained embeddings and fine-tuned task specific models
  \item \cpberturl\ pseudo training dataset generation and model source code
    used in our experiments
  \end{zlenumerateinline}
\end{nohyphenation}
\end{abstract}


\acresetall 

\sampleFirstName[ht!][\columnwidth]


\zssec{Introduction}

De-identification is the process of modifying a corpus so that any personal or
private information is either redacted (masked data), replaced with from the
same corpus (anonymized data), or replaced from an external source (synthetic
data).  An example of a de-identified dataset is \mimiccite, which uses
\acp{maskt} that have a particular syntax and identify the masked/redacted
text.  For example, \masktex\ are \acp{maskt} that replace a physician's last
name as shown in \zfrefsub{sample-first-name}{b}.  In this work, ``masked''
tokens or text refer to de-identified redacted text rather than mask tokens in
\mtms\ such as \bert~\cite{\bertct}.  This clarification becomes more important
in the discussion of the mask language pretrained embeddings in
\zssecref{meth:pttrain}.

Clinical data is governed by \ac{hipaa} laws, which protect patients' health
records.  De-identified corpora do exist, such as the \acl{crwd}~\cite{\crwdct}
and \chf, but these datasets take significant effort to procure.  However,
\mimic\ is publicly available and easily accessible\footnote{Access to the
  \mimic\ corpus \href{https://mimic.mit.edu/docs/gettingstarted/}{requires}
  creating a PhysioNet account and finishing a training course.}, which is the
reason it is the most popular corpus for clinical text research.  \mimic\ has
been used to pretrain many \llms\ and
\mtms~\cite{yangGatorTronLargeClinical2022,huangClinicalBERTModelingClinical2020,alsentzerPubliclyAvailableClinical2019}.

Given the popularity of \mimic\ and that a significant portion of the corpus is
redacted (11,204,493 \acp{maskt} or 6\% of the corpus), we believe it warrants
investigation to determine the negative effects of redacted mask tokens.  Our
primary research questions for this work are
\begin{zlenumerateinline}
\item to determine if \mtms\ are negatively effected by redacted text
\item if they are, can they be improved by pretraining on substituted realistic
  pseudo data
\end{zlenumerateinline}
We hypothesize that named entities in language models lack the necessary
entropy, and instead, saturate over the redacted tokens in de-identified
corpora.  We believe that other tasks, such as \ner\ and entity linking, are
also negatively affected for the same reasons.

To address these questions, we processed \mimic's entire two million free text
notes written by medical professionals.  These clinical notes were processed
for unsupervised pretraining with masked text left intact
(see~\zfrefsub{sample-first-name}{b}); we call this the ``\mds''.  This corpus
was copied with masked text replaced by pseudo text, which resulted in the
``\pds'' (see~\zfrefsub{sample-first-name}{c}).  We then used the original
\mds\ and \pds\ to self-supervised train several encoder-only \mtms\ and save
each as new pretrained checkpoints.  Each checkpoint was then used to train
and evaluate new fine-tuned models on a de-identification dataset.

The pseudo data generation techniques are covered in \zssecref{meth:pseudodb},
and the methods to create the \pds\ are covered in \zssecref{meth:datasets}.
The de-identification models, architecture and method of training are detailed
in \zssecref{expsetup}.  We explain how our pretraining methods led to better
fine-tuned de-identification models with a surprising finding in
\zssecref{results} and discuss trade-offs in~\zssecref{disc}.


\zssec{Related Work}


As the \ac{sota} of general domain \llms\ outpaced that of clinical
\llms~\cite{deepseek-aiDeepSeekR1IncentivizingReasoning2025,anilGeminiFamilyHighly2024,touvronLLaMAOpenEfficient2023},
their necessity was
challenged~\cite{agrawalLargeLanguageModels2022,kartchnerZeroShotInformationExtraction2023}.
However, recent studies show that clinical domain \llms\ and \mtms\ have better
performance and are more efficient at biomedical domain specific
tasks~\cite{wangDRGLLaMATuningLLaMA2024,hagerEvaluationMitigationLimitations2024,OpenBioLLMs,wornowShakyFoundationsClinical2023,lehmanWeStillNeed2023}.


Others have used pseudo tokens in pretrained language models.
\zeciteshort{tanSentenceWorth1282022} used pseudo tokens to train new BERT
layers for contrastive learning.  However, they use a generative method for
semantically similar sentences where ours uses classification models to
determine pretrained embedding models' performance on a biomedical task.
Recently, \llms\ have been leveraged to produce pseudo data, such as rare
molecule
discovery~\cite{chenArtificiallyRealReal2024,guoMolTailorTailoringChemical2024}
and open-ended responses
generation~\cite{zenimotoCodingOpenEndedResponses2024}.



The de-identification of clinical corpora using contemporary \ac{dl} methods
started with the work of
\zeciteshort{liuAutomaticDeidentificationElectronic2015}, who used character
level \acp{nn}, and \zeciteshort{\derndidct}, who used \ac{bilstm}\ with
non-contextual word
embeddings~\cite{mikolovDistributedRepresentationsWords2013,mikolovEfficientEstimationWord2013,penningtonGloveGlobalVectors2014}.
Similarly, \zeciteshort{\liuctdeid} used a \ac{bilstm} with a \ac{crf} output,
and \zeciteshort{\hartmanct} used an \ac{rnn} with a \ac{crf} to classify
\ac{hphi}.  Soon after, \zeciteshort{\johndidct} utilized \mtms, such as \bert\
for the task.

We also use \mtms, but train our own checkpoints rather than use off-the-shelf
models.  Our goal differs in that we endeavor to discover how check-pointing
pretrained embeddings on masked and pseudo tokens affect the downstream
fine-tuning task of de-identification.


\zssec[dataset]{Dataset}

This work involves two types of datasets: \mtm\ (also suitable for \llms)
pretraining datasets and a preexisting de-identification dataset with \ac{hphi}
annotations used for fine-tuning models for evaluation.  The pretraining
datasets include \ztsee{gencorptab}:
\begin{zlpackeditemize}
\item \textbf{Masked}: Formatted clinical notes with \acp{maskt} in place of
  redacted \ac{hphi}.
\item \textbf{Pseudo}: A copy of the \mds\ with \acp{maskt} replaced with
  pseudo data.
\end{zlpackeditemize}

\gencorptab[t]{\tabsize}

\zspara[dataset:pt]{Pretrained}

Both the \mdp\ are derived from \mimiccite.  This corpus is taken from the Beth
Israel Deaconess Medical Center in Boston, Massachusetts and contains 58,976
hospital admissions.  This dataset includes both structured (such as patient
vitals) and unstructured data (such as free text notes by medical
professionals).  Our datasets are based on the \mimicname\ unstructured data
and consists of 2,083,180 notes from 46,520 patients who were admitted to the
\ac{icu} surgical, medical, and neonatal departments.

\zspara[dataset:deid]{De-identification}

The \ac{itbtdeid} de-identification dataset~\cite{\itbtdeidct} that reports on 296
patients in 1,304 medical records.  This dataset was de-anonymized by replacing
\ac{hphi} with hand crafted pseudo data, which is somewhat similar to our method.
The redacted data was then demarcated with lexical spans annotated with one of
23 labels indicating the type of replaced data, such as ``date'', ``hospital'',
etc.


\zssec[meth]{Methodology}

Our methods are split in to two phases: pretraining the embeddings and
fine-tuning the de-identification model.  \zfref{sample-first-name} shows
the end-to-end sequence of steps of our method.  The process overview follows:
\begin{zlpackedenum}
\item Create the pseudo database with gazetteer like data
  lists~(\zfrefsub{sample-first-name}{a}, \zssecref{meth:pseudodb})%
  \label{meth:pseudodb:step}
\item Create the derived \mds~(\zfrefsub{sample-first-name}{b}) and
  \pds~(\zfrefsub{sample-first-name}{c}, \zssecref{meth:datasets}).
\item Train embeddings~(\zfrefsub{sample-first-name}{d},
  \zssecref{meth:pttrain}).
\item Fine-tune models trained on the de-identification corpus using the masked
  and pseudo checkpoints~(\zfrefsub{sample-first-name}{c},
  \zssecref{expsetup}).
\item Evaluate the fine-tuned model~(\zssecref{results}).
\end{zlpackedenum}

\genmetatab[t]{\smalltabsize}

\zssubsec[meth:pretrain]{Pretraining}

The pretraining phase included creating the \mdp, and then training embeddings.

\zssubsubsec[meth:pseudodb]{Pseudo Database}

The first preprocessing step was to create a pseudo database.  This database
includes lists of data that is used to derive the \pds.  These lists and their
counts include:
\begin{zlitemizeinline}
\item Hospitals by state (4,826)
\item Companies (500)
\item Universities (2,074)
\item US states names and abbreviations (51)
\item First names by gender (28,516)
\item Last names (162,252)
\end{zlitemizeinline}

The first name, gender and popularity distribution comes from a collection of
sources.  First and last names include popularity statistics as a count
recorded by year.  The last name popularity distribution comes from the 2010 US
census.  The first name popularity distribution was drawn from the years
between 1960 and 2020 to better represent those that would be found in the
\mimicname\ corpus.


\zssubsubsec[meth:datasets]{Derived Datasets}

The \mds\ was created by formatting the clinical text as a single flat newline
separated file with normalized sentences.  This was accomplished by tokenizing
and sentence chunking the contents of the free text clinical
\texttt{NOTEEVENTS} table from \mimic~\zsseesec{dataset:pt} using
ScispaCy~\cite{neumannScispaCyFastRobust2019} library.  We also reconstructed
enumerated and itemized lists (\zeie\ vitals) heuristically.

The \pds\ was then created by replacing the \acp{maskt} with realistic text by
sampling from the pseudo database~\zsseesec{meth:pseudodb} and randomized data
generation methods.  For each line of the \mds, tokens were parsed for
\ac{hphi} ``tags'' with \mimicname\ mask regular expressions.  The tag (such as
a first name, hospital, medical record number, etc.), was then used to select
how the pseudo data was generated for the replacement.  For example, ``Doctor
First Name'' would be parsed from \masktex\ using a regular expression, which
in turn was mapped to the tag \texttt{DOCTORLASTNAME}.  A list of data
generators and the tag masked token tags associated with them is given in
\genmetaref.

Once a generator was selected, it was used to create the pseudo text.  For
example, the last name generator mapped from the \texttt{DOCTORLASTNAME} tag
would sample the pseudo database for the replacement text, and substitute it in
place of the \mimicname\ \acp{maskt}. The pseudo data replacement of first and
last names were sampled from the database using the popularity of the name as
the probability distribution.  Some tags include the gender of the name, or
chosen randomly if the tag was missing.

Much of the generated data comes from randomized data rather than lists.
Examples include dates, identifiers such as medical record numbers
(\texttt{NUMERICID}) and phone numbers (\texttt{PHONE}), and patient ages
(\texttt{AGE}).

Not all \mimic\ \acp{maskt} have a generator as the text often does not follow a
conventional surface text pattern.  In these cases we use the clinical \llm\
GatorTron~\cite{yangGatorTronLargeClinical2022} to fill in the missing token
using the entire sentence as input.  The model was also used to replace
ambiguous tags such as \texttt{NAME}, which in \mimicname, can name a person,
place or thing.

\zssubsubsec[meth:pttrain]{Pretraining}

\checkPointHistory[b!]

Once the \mdp\ were created, the language models were trained from the
checkpoints from two popular models: \roberta~\cite{\robertact} and
\xlmroberta~\cite{\xlmrobertact}.  The base and large versions of these models
were trained on the masked and pseudo training sets from their last
checkpoints, which resulted in eight pretrained embedding checkpoints.  These
model types and sizes were chosen for their diversity and because they are
masked-trained language models on \dsp\ that did not include clinical or
biomedical text.

Initially, the \roberta\ checkpoint was used to train on the \mds\ for
hyperparameter tuning on the fine-tuned de-identification model
\zsseesec{expsetup} specifically for optimizing the embeddings.  We chose the
pretraining model hyperparameter set by training four embedding models and then
fine-tuning a model for each embedding.  The hyperparameter set for the highest
performing fine-tuned model was then fixed and used to train all pretrained
embedding models for six epochs.

\pretrainHyperparamTab[t]{\tabsize}

The fixed hyperparameters include
\begin{zlitemizeinline}
\item a learning rate of $2 \times 10^{-5}$
\item max sequence length of 1024
\item the mini-batch sizes~\footnote{The mini-batch size was based on GPU
    memory constraints.} in~\ztref{pretrainHyperparamTab}
\end{zlitemizeinline}

All pretrained embeddings were trained on a 2.8 GHz AMD EPYC Milan 32 core CPU
with 512 GB CPU RAM distributed across four A100 GPUs with a combined 160 GB of
video RAM\@.  Training time varied for each model, but the larger models took
between 39 and 40 hours.

After the embedding models were trained, fine-tuned models were trained from
each of their check points as shown in \zfref{checkpoint-history}.


\zssec[expsetup]{Experimental Setup}

\labelDistNonMappedTab[t]{\tabsize}

Our hypothesis is that language models learn the masks used in redacted data
(such as \masktex).  To test this hypothesis, we used fill-mask prediction of
the mask and pseudo \roberta\ large models and compared the results.  Text
was taken directly from \mimicname\ and used as input to the models with the
redacted masks replaced with the fill-mask token (\texttt{[MASK]}).  While this
illuminates properties of the embeddings, we also wanted measure the impact of
masked tokens on task-specific fine-tuned models.

To accomplish this, we fine-tuned eight models on the
\ac{itbtdeid}~\cite{\itbtdeidct} de-identification dataset.  The task was
framed as named entity recognition by classifying \ac{hphi} labeled tokens
using the same training and test splits from prior
work~\cite{\derndidct,\johndidct}.  However, our validation set was created
from 20\% of the training set's clinical notes\footnote{A validation set has
  not yet been formalized.}  using multi-label iterative
stratification~\cite{sechidisStratificationMultilabelData2011}.  See
\ztref{labelDistNonMappedTab} for the label dataset split distributions.

\examplePretrainOutputTab[t]{\tabsize}

\labelDistMappedTab[b]{\tabsize}

We trained and evaluated the fine-tuned models on the \itbtname\
de-identification dataset labels.  However, we re-categorized the test set gold
labels and predictions to the \ac{hipaa} classification used
by~\zeciteshort{\derndidct} to compare with previous
results~\zsseesec{results}.  Even though we suspected that training on the
larger \itbtname\ label set could negatively impact results, the original
labels were used to provide additional data points for comparison between the
masked and pseudo embeddings.  See \ztref{labelDistMappedTab} for the label
re-categorization dataset split distributions.

The fine-tuned de-identification models were trained from each checkpoint of
the eight pretrained embedding models described in~\zsseesec{meth:pttrain}.  A
fully connected linear layer was added between the transformer's final layer
and the output layer.  All models were trained with the same hyperparameters,
which included a learning rate of $7 \times 10^{-6}$, a drop out of 10\%, and
trained for 20 epochs using the model with the lowest validation loss.  The
validation loss in \zfref{fine-tuned-model-loss} shows the larger models
converging sooner than the base models as expected.

In addition to our masked and pseudo embeddings, we fine-tuned models from the
checkpoints of several baseline models to provide additional points of
reference.  These embeddings include the \bert~\cite{\bertct} uncased base
and small, \roberta\ base and \bioclinicbert~\cite{\bioclinicbertct}.  The
baseline models were trained using the same de-identification dataset,
hyperparameters and procedure as the masked and pseudo fine-tuned models.

\ifisfinal%
The Zensols Deep NLP framework~\cite{landesDeepZensolsDeepLearning2023} was
used for fine-tuning model development, training, and evaluation.
\fi

\fineTuneDeidLoss[b!]{0.9\columnwidth}


\zssec[results]{Results}

We report experimental findings of the pretrained embeddings and the evaluation
results on the fine-tuned de-identification models.

\zspara[res:pt]{Pretrained Embeddings}

With little effort, we were able to induce \mimicname\ masked tokens from the
\roberta\ large masked pretrained embeddings as described in
\zssecref{expsetup}.  The masked model output is listed in
\ztref{examplePretrainOutputTab}, and includes masked tokens such as
asterisks (\texttt{*}) and brackets (\texttt{[}, \texttt{]}).

On the other hand, the output of the pseudo model produces realistic clinical
text that would be written by a physician.  For example, the replaced token in
``Delivering OB : Dr. \texttt{[MASK]} Pediatrician'' is an asterisk in the
masked pretrained embedding model, but the pseudo model produces the last name
``Jones''.

\zspara[res:ft]{Fine-tuned De-identification}

\finetuneModelResultsTab[t]{\tabsize}

The fine-tuned results given in \ztref{finetuneModelResultsTab}.  When
discussing these models, we refer to them by their index in the first column.
Our baseline models (6 -- 9), which were trained on de-identification dataset,
show an improvement over previous work's baseline (1 -- 5), which highlight our
fine-tuned model fitness in architecture and/or hyperparameter set.  The lower
performance of \bioclinicbert~\cite{\bioclinicbertct} compared to all of our
masked models is notable as it was also trained on \mimicname.  We believe this
is some combination of our preprocessing of the \mds~\zsseesec{meth:datasets},
or more likely the hyperparameter tuning of the masked
model~\zsseesec{meth:pttrain}.

Our masked and pseudo models (10 --- 17) perform significantly better than all
previous baselines and off-the-shelf pretrained embeddings baseline models.
Our highest performing model (12) surpasses the best baseline model (5) by 1.5
F1 points~\cite{\johndidct}, which is a large margin given the already high
performance of the task.  Surprisingly, this masked model outperformed its peer
pseudo model~\zsseesec{disc}.  The difference in performance between the
\roberta\ base models (10, 11) were shown not to be statistically
significant using McNemar's test~\cite{mcnemarNoteSamplingError1947} with
$p_m=0.7$ and a one-way ANOVA having $p_a=0.8$.  However, the results are
significantly different using the same statistics on the remaining models all
having a $p_m < 6.4 \times 10^{-3}$ and $p_a < 0.035$.


\finetuneModelResultsByLabelTab[t]{\tabsize}

\zssec[disc]{Discussion and Error Analysis}

\fineTuneConfusion[b!]{\columnwidth,height=13.3cm}

Our hypothesis that pseudo models outperform masked models holds for
\xlmroberta\ but not \roberta\ large.
\zfref{fine-tuned-confusion-mapped} shows that the pseudo model incorrectly
predicted \texttt{LOCATION} 36 times more than the masked model.  These
(re-categorized \texttt{LOCATION}) false positives are attributed to 34
additional \itbtname\ corpus non-\ac{hphi} (``other'' \texttt{O}) tokens categorized as
\ifisfinal%
  \texttt{HOSPITAL} (see \zsapxref{conf-non-mapped}
  \zfref{fine-tuned-confusion-non-mapped-rob-large}).
\else
  \texttt{HOSPITAL}.
\fi

Many of the misclassified locations include named entities and medical terms
such as Medtronic (medical device company) and TSH (thyroid stimulating
hormone).  Generally, the model performed better with context words that helped
in clarifying the terminology such as ``clinic'' and ``north'' in hospital
names.

Given the high number of \texttt{LOCATION} misclassifications we calculated the
intersection of hospital names in the \itbtname\ dataset with the pseudo list
and found that only 29 (7.2\%) are shared, or 35 (8.91\%) if ``hospital'' and
``clinic'' are stripped.  Yet, this tag was the second most replaced data type
in the \pds\ (see \genmetaref).

We see a similar pattern with non-\ac{hphi} tokens predicted as \texttt{NAME} 24
times more often in the pseudo model.  We also see that \texttt{NAME} is
mislabeled as \texttt{LOCATION} 23 more times in the pseudo model.  The
majority of the misclassifications are the \itbtname\ \texttt{PATIENT} false positives,
which are often confused with doctor names.  This speaks to a larger problem of
confusion with medical terms and human names.

However, the pseudo model was better with diseases and medical terms named after
individuals.  For example, a mention of the last name ``Gleason'', which as a
prostate cancer pathology scoring system~\cite{gleason1992histologic}, was
correctly labeled in the pseudo model but not the masked.


The performance results in shown in \ztref{finetuneModelResultsByLabelTab}
support the these findings.  \texttt{NAME} is shown to under perform in the
\roberta\ pseudo models except for the \xlmroberta\ models.  In fact, we see
that the \xlmroberta\ pseudo models consistently outperform the masked models
\ifisfinal%
  (with the exception of the \texttt{AGE} label). See models 13, 14 in
  \zsapxref{conf-non-mapped} \zfref{fine-tuned-confusion-non-mapped-rob-large}
  for confusion matrices.
\fi

This is partly due to the sensitivity of the score given the low count of the
label (824 in the training set).  Another factor is almost certainly the single
numeric-only values replaced in the \pds, such as \texttt{AGE} tags (see
\genmetaref).  Furthermore, the \itbtname\ corpus includes plural morphology
(``40's'') other, descriptors such as years and month (``49y7.7m''), which were
not addressed when constructing the pseudo data generators.

The high performance, despite potentially low entropy or confusing pseudo data,
highlights the robustness of the \xlmroberta\ model.  We see a significant
improvement in the pseudo over the masked models for \texttt{PROFESSION},
\texttt{NAME} and \texttt{LOCATION}.  We believe the \xlmroberta\ model is more
capable with these labels and Latin-rooted medical terminology because of its
exposure to foreign names, locations and romance languages.



\zssec[conclusion]{Conclusion and Future Work}

Our hypothesis that pseudo dataset \mtm\ training yields fitter downstream
fine-tuned de-identification models is congruent with the multi-lingual
\xlmroberta.  Multi-lingual trained \mtms\ are better adept at continued
learning on lower entropy pseudo data (such as foreign names).  However, the
quality of generated pseudo data can negatively impact downstream fine-tuned
model performance.  Furthermore, there is no one-size-fits-all for all tasks as
any \mimic\ trained model will generate masked tokens.  However, clinical
models still have a place as they, thus far, continue to outperform general
domain foundational
models~\cite{huInformationExtractionClinical2024,wornowShakyFoundationsClinical2023}.

The pseudo data trained models improved results and implies real clinical
training would increase performance even further for the de-identification
task.
Subsequent work includes self-supervised training \llms\ on our datasets and
supervised fine-tuning new clinical task models from the pretrained
checkpoints.


\zavenenddoc{cpbert}

\ifisfinal%
\clearpage
\onecolumn
\zavenappendix%





\zsapx[conf-non-mapped]{De-identification \itbtname\ \ac{hipaa} Label Confusion Matrices}

\fineTuneConfusionNonMappedRobSmall[h!]

\fineTuneConfusionNonMappedRobLarge[h!]

\fineTuneConfusionNonMappedRobXlmSmall[h!]

\fineTuneConfusionNonMappedRobXlmLarge[h!]

\clearpage

\zsapx[meth:pseudodb:list]{Pseudo Database Lists}

\pseudosrctab[h]{\tabsize}

\zsapx[limitations]{Limitations}

While a variety of baselines were used for comparison and the \ac{itbtdeid} corpus
is sufficiently large, other combinations of embeddings and fine-tuned tasks
could yield different results.

\fi

\end{document}